%% file: emnlp2020.tex
\documentclass[11pt,a4paper]{article}
\usepackage[hyperref]{emnlp2020}
\usepackage{times}
\usepackage{latexsym}

\usepackage{textcomp}

\input{math_commands.tex}

\usepackage{microtype}

\aclfinalcopy 

\usepackage{multirow}

\usepackage{amssymb}
\usepackage{graphicx}
\usepackage{subcaption}
\usepackage{tabularx} 

\usepackage{ctable}

\newcommand{\Expect}{{\rm I\kern-.3em E}}

\newcommand{\x}{\boldsymbol{x}}
\newcommand{\y}{\boldsymbol{y}}
\newcommand{\xspace}{\mathcal{X}}
\newcommand{\yspace}{\mathcal{Y}}
\newcommand{\relyspace}{\mathcal{Y}_{R}}
\newcommand{\cost}{\bigtriangleup}
\newcommand{\infnet}{\mathbf{A}_{\Psi}}
\newcommand{\canet}{\mathbf{F}_{\Phi}}

\newenvironment{itemizesquish}{\begin{list}{\labelitemi}{\setlength{\itemsep}{-0.2em}\setlength{\labelwidth}{0.5em}\setlength{\leftmargin}{\labelwidth}\addtolength{\leftmargin}{\labelsep}}}{\end{list}}

\title{Improving Joint Training of Inference Networks and \\ Structured Prediction Energy Networks}

\author{Lifu Tu$^1$ \ \ \ \ \ \ \ Richard Yuanzhe Pang$^2$\thanks{~~Work done at the University of Chicago and Toyota Technological Institute at Chicago.} \ \ \ \ \ \ \ Kevin Gimpel$^1$ \\
$^1$Toyota Technological Institute at Chicago, Chicago, IL 60637, USA \\
$^2$New York University, New York, NY 10011, USA \\
{\tt lifu@ttic.edu, yzpang@nyu.edu, kgimpel@ttic.edu}}

\date{}

\begin{document}
\maketitle
\begin{abstract}
 Deep energy-based models are powerful, but pose challenges for learning and inference~\citep{belanger2016structured}. 
  \citet{tu-18} developed an efficient framework for energy-based models by training ``inference networks'' to approximate structured inference instead of using gradient descent. 
  However, their alternating optimization approach suffers from instabilities during training, requiring additional loss terms and careful hyperparameter tuning. In this paper, we contribute several strategies to stabilize and improve this joint training of energy functions and inference networks for structured prediction. We design a compound objective to jointly train both cost-augmented and test-time inference networks along with the energy function. We propose joint 
  parameterizations for the inference networks that encourage them to capture complementary functionality during learning. We empirically validate our strategies on two sequence labeling tasks, showing easier paths to strong performance than prior work, as well as further improvements with global energy terms.
\end{abstract}

\input{intro}

\input{background.tex}

\input{methods}

\input{new_tech}

\input{energy}

\input{ExperimentSetup.tex}

\input{experiments}

\input{relatedwork}

\section{Conclusions and Future Work}

We contributed several strategies to stabilize and improve joint training of SPENs and inference networks. Our use of
joint parameterizations mitigates the need for inference network fine-tuning, leads to complementarity in the learned inference networks, 
and yields improved performance overall. These developments offer promise for SPENs to be more easily applied to a broad 
range of NLP tasks. 
Future work will explore other structured prediction tasks, such as parsing and generation. 
We have taken initial steps in this direction, considering constituency parsing with the 
sequence-to-sequence model of \citet{tran-etal-2018-parsing}. 
Preliminary experiments are positive,\footnote{On NXT Switchboard \citep{calhoun2010nxt}, the baseline achieves 82.80 F1 on the development set and the SPEN (stacked parameterization) achieves 83.22. More details are in the appendix.} 
but significant challenges remain, specifically in defining appropriate inference network architectures to enable efficient learning.

\section*{Acknowledgments}
We would like to thank the reviewers for insightful comments. 
This research was supported in part by an Amazon Research Award to K.~Gimpel.

\bibliographystyle{acl_natbib}
\bibliography{anthology,emnlp2020}

\appendix

\section{Appendices}
\label{sec:appendix}

\subsection{Constituency Parsing Experiments}

We linearize the constituency parsing outputs, similar to \citet{tran-etal-2018-parsing}. We use the following equation plus global energy in the form of Eq.~(8) as the energy function: 
\begin{align}
\label{eqn:appendix-energy-sequence-labeling}
E_{\Theta}(\x, \y) &= -\Bigg(\sum_{t=1}^T  \sum_{j=1}^L y_{t,j}\! \left(U_j^\top b(\x,t)\right) \nonumber \\
& + \sum_{t=1}^T \y_{t-1}^\top W \y_{t}\Bigg) \nonumber 
\end{align}
Here, $b$ has a seq2seq-with-attention architecture identical to \citet{tran-etal-2018-parsing}. In particular, here is the list of implementation decisions. 
\begin{itemizesquish}
    \item We can write $b=g\circ f$ where $f$ (which we call the ``feature network'') takes in an input sentence, passes it through the encoder, and passes the encoder output to the decoder feature layer to obtain hidden states; $g$ takes in the hidden states and passes them into the rest of the layers in the decoder. In our experiments, the cost-augmented inference network $\canet$, test-time inference network $\infnet$, and $b$ of the energy function above share the same feature network (defined as $f$ above). 
    \item The feature network ($f$) component of $b$ is pretrained using the feed-forward local cross-entropy objective. The cost-augmented inference network $\canet$ and the test-time inference network $\infnet$ are both pretrained using the feed-forward local cross-entropy objective. 
\end{itemizesquish}

The seq2seq baseline achieves 82.80 F1 on the development set in our replication of \citet{tran-etal-2018-parsing}. Using a SPEN with our stacked parameterization, we obtain 83.22 F1. 


\end{document}

%% file: math_commands.tex
\usepackage{amsmath,amsfonts,bm}

\def\eqref#1{equation~(\ref{#1})}

\def\1{\bm{1}}

\def\vx{{\bm{x}}}
\def\vy{{\bm{y}}}

\DeclareMathAlphabet{\mathsfit}{\encodingdefault}{\sfdefault}{m}{sl}
\SetMathAlphabet{\mathsfit}{bold}{\encodingdefault}{\sfdefault}{bx}{n}

\newcommand{\softmax}{\mathrm{softmax}}

\DeclareMathOperator*{\argmax}{arg\,max}
\DeclareMathOperator*{\argmin}{arg\,min}

%% file: intro.tex
\section{Introduction}

Energy-based modeling \citep{lecun-06} associates a scalar compatibility measure to each configuration of input and output variables. 
\citet{belanger2016structured} formulated deep energy-based models for structured prediction, which they called structured prediction energy networks (SPENs). 
SPENs use arbitrary neural networks to define the scoring function over input/output pairs. However, this flexibility leads to challenges for learning and inference. The original work on SPENs 
used gradient descent for structured inference~\citep{belanger2016structured,End-to-EndSPEN}. 
\citet{tu-18,tu-gimpel-2019-benchmarking} found improvements in both speed and accuracy by replacing the use of gradient descent with a method that trains a neural network (called an ``inference network'') to do inference directly. 
Their formulation, which jointly trains the inference network and energy function, is similar to training in generative adversarial networks~\citep{goodfellow2014generative}, which is known to suffer from practical difficulties in training due to the use of alternating optimization~\citep{NIPS2016_6125}. 
To stabilize training, \citet{tu-18} experimented with  several additional terms in the training objectives, finding performance to be dependent on their inclusion. 

Moreover, when using the approach of \citet{tu-18}, there is a mismatch between the training and test-time uses of the trained inference network. During training with hinge loss, the inference network is actually trained to do ``cost-augmented'' inference. However, at test time, the goal is to simply minimize the energy without any cost term. \citet{tu-18} fine-tuned the cost-augmented network to match the test-time criterion, but found only minimal change from this fine-tuning. 
This suggests that the cost-augmented network was mostly acting as a test-time inference network by  convergence, which may be hindering the potential contributions of cost-augmented inference in max-margin structured learning~\citep{ssvm,m3}.

In this paper, we contribute a new training objective for SPENs that addresses the above concern and also contribute several techniques for stabilizing and improving learning. 
We empirically validate our strategies on two sequence labeling tasks from natural language processing (NLP), namely part-of-speech tagging and named entity recognition. We show easier paths to strong performance than prior work, as well as further improvements with global energy terms.
We summarize our list of contributions as follows. 
\begin{itemize}
    \item We design a compound objective under the SPEN framework to jointly train the ``training-time'' cost-augmented inference network and test-time inference network (Section~\ref{sec:joint}).
    \item  We propose shared parameterizations for the two inference networks so as to encourage them to capture complementary functionality while reducing the total number of trained parameters (Section~\ref{sec:joint-param}). Quantitative and qualitative analysis shows clear differences in the characteristics of the two networks (Table~\ref{table:phi-vs-psi}).
    \item We present three methods to streamline and stabilize training that help with both the old and new objectives (Section~\ref{sec:techniques}).
    \item We propose global energy terms to capture long-distance dependencies and obtain further improvements (Section~\ref{sec:tlm-training}).
\end{itemize}

\noindent While SPENs have been used for multiple NLP tasks, including multi-label classification~\citep{belanger2016structured}, part-of-speech tagging~\citep{tu-18}, and semantic role labeling~\citep{End-to-EndSPEN}, they are not widely used in NLP. Structured prediction is extremely common in NLP, but is typically approached using methods that are more limited than SPENs (such as conditional random fields) or models that suffer from a train/test mismatch (such as most auto-regressive models). SPENs offer a maximally expressive framework for structured prediction while avoiding the train/test mismatch and therefore offer great potential for NLP. However, the training and inference have deterred NLP researchers. 
While we have found benefit from training inference networks for machine translation in recent work \citep{tu-etal-2020-engine}, that work assumed a fixed, pretrained energy function. 
Our hope is that the methods in this paper will enable SPENs to be applied to a larger set of applications, including generation tasks in the future. 

%% file: background.tex
\section{Background} 
\label{sec:background}


We denote the input space by $\xspace$. For an input $\x\in\xspace$, we denote the structured  output space by $\yspace(\x)$. The entire space of structured outputs is denoted $\yspace = \cup_{\x\in\xspace} \yspace(\x)$. 
A SPEN~\citep{belanger2016structured} defines an \textbf{energy function} $E_{\Theta} : \xspace \times \yspace \rightarrow \mathbb{R}$ parameterized by $\Theta$ that 
computes a scalar energy for an input/output pair. 
At test time, for a given input $\x$, prediction is done by choosing the output with lowest energy:
\begin{align}\label{eq:inf}
\textstyle{\hat{\y} = \argmin_{\y\in\yspace(\x)}E_{\Theta}(\x, \y)}
\end{align}
However, solving \eqref{eq:inf} requires combinatorial algorithms because $\yspace$ is a structured, discrete space. This becomes intractable when $E_{\Theta}$ does not decompose into a sum over small ``parts'' of $\y$. 
\citet{belanger2016structured} relaxed this problem by allowing the \textbf{discrete vector} $\y$ to be continuous; $\relyspace$ denotes the \textbf{relaxed output space}. 
They solved the relaxed problem by using gradient descent to iteratively minimize the energy with respect to $\y$. The energy function parameters $\Theta$ are trained using a structured hinge loss which requires repeated cost-augmented inference during training. 
Using gradient descent for the repeated cost-augmented inference steps is time-consuming and makes learning unstable~\citep{End-to-EndSPEN}. 

\citet{tu-18} 
replaced gradient descent  
with a neural network trained to do efficient inference. This ``inference network'' $\infnet : \xspace \rightarrow \relyspace$ is parameterized by $\Psi$ and trained with the goal that
\begin{align}\label{eqn:original-psi}
\infnet(\x) \approx \argmin_{\y\in\relyspace(\x)}E_\Theta(\x, \y)
\end{align}
When training the energy function parameters~$\Theta$, \citet{tu-18} replaced the cost-augmented inference step in the structured hinge loss from \citet{belanger2016structured} 
with a cost-augmented inference network 
$\canet$:
\begin{align}\label{eqn:costaug}
\canet(\x) \approx \argmin_{\y\in\relyspace(\x)}\,  (E_\Theta(\x, \y)-\cost(\y, \y^\ast))
\end{align}
\noindent where $\cost$ is a structured cost function that computes the distance between its two arguments. We use
L1 distance for $\cost$. This inference problem involves finding an output with low energy but high cost relative to the gold standard. Thus, it is not well-aligned with the test-time inference problem.

Here is the specific objective to jointly train $\Theta$ (parameters of the energy function) and $\Phi$ (parameters of the cost-augmented inference network): 
\begin{align}
\label{eqn:original-training}
\small
\min_{\Theta}& \max_{\Phi} \sum_{\langle \x_i, \y_i\rangle\in\mathcal{D}}  [ \cost(\canet(\x_i), \y_i) \nonumber \\ & -  E_{\Theta}(\x_i,\canet(\x_i))  + E_{\Theta}(\x_i, \y_i) ]_{+}
\end{align}
\noindent where $\mathcal{D}$ is the set of training pairs, $[h]_+ = \max(0,h)$, and $\cost$ is a structured cost function that computes the distance between its two arguments. 
\citet{tu-18} 
alternatively optimized $\Theta$ and $\Phi$, which is similar to training in generative adversarial networks~\citep{goodfellow2014generative}. The inference network is analogous to the generator and the energy function is analogous to the discriminator.
As alternating optimization can be difficult in practice~\citep{NIPS2016_6125}, 
Tu \& Gimpel experimented with including several additional terms in the above objective to stabilize training. 

After the training of the energy function, an inference network $\infnet$ for test-time prediction is fine-tuned with the goal shown in Eq.~(\ref{eqn:original-psi}). More specifically, for the fine-tuning step, we first initialize $\Psi$ with $\Phi$; next, we do gradient descent according to the following objective to learn $\Psi$:
$$\Psi \leftarrow \argmin_{\Psi'} \sum_{\x \in\mathcal{X}} E_\Theta (\vx, A_{\Psi'} (\vx))$$
where $\mathcal{X}$ is a set of training or validation inputs. It could also be the test inputs in a transductive setting.




%% file: methods.tex
\section{An Objective for Joint Learning of Inference Networks} 
\label{sec:joint}
\label{sec:objective}

One challenge with the above optimization problem is that it requires training a separate inference network $\infnet$ for test-time prediction after the energy function is trained. 
In this paper, we propose an alternative that trains the energy function and both inference networks jointly. In particular, we use a ``compound'' objective that combines two widely-used losses in structured prediction. We first present it without inference networks:
\begin{align}
\min_{\Theta} \!\!
&\sum_{\langle \x_i, \y_i\rangle\in\mathcal{D}} \nonumber \\
& \!\!\!\!\!\!\underbrace{\left[\max_{\y}  (\cost(\y, \y_i) \!-\! E_{\Theta}(\x_i,\y) \!+\! E_{\Theta}(\x_i, \y_i))\!\right]_{\! +}}_{\text{margin-rescaled hinge loss}} \nonumber \\
& \!\!\!\!\!\! + \lambda\underbrace{\left[ \max_{\y}( - E_{\Theta}(\x_i,\y) + E_{\Theta}(\x_i, \y_i))\right]_{\! +}}_{\text{perceptron loss}} 
\end{align}
\noindent As indicated, this loss can be viewed as the sum of the margin-rescaled hinge and perceptron losses for SPENs. 
Two different inference problems are represented. 
The margin-rescaled hinge loss contains cost-augmented inference, shown as part of Eq.~(\ref{eqn:costaug}). 
The perceptron loss contains the test-time inference problem, which is shown in Eq.~(\ref{eq:inf}). 
\citet{tu-18} used a single inference network for solving both problems, so it was trained as a cost-augmented inference network during training and then fine-tuned as a test-time inference network afterward. 
We avoid this issue by training two inference networks, $\infnet$ for test-time inference and $\canet$ for cost-augmented inference:
\begin{align}
& \!\min_{\Theta} \max_{\Phi, \Psi} \!\!\!\!\sum_{\langle \x_i, \y_i\rangle\in\mathcal{D}} \nonumber \\ 
& 
[\cost(\canet(\x_i), \y_i) \!-\! E_{\Theta}(\x_i,\canet(\x_i)) \!+\! E_{\Theta}(\x_i, \y_i)]_{+}
\nonumber \\
& + 
\lambda 
\left[- E_{\Theta}(\x_i,\infnet(\x_i)) \!+\! E_{\Theta}(\x_i, \y_i)\right]_{+}
\label{eq:margin-hinge} 
\end{align}
We treat this optimization problem as a minimax game and find a saddle point for the game similar to \citet{tu-18} and \citet{goodfellow2014generative}. We use minibatch stochastic gradient descent and alternately optimize $\Theta$, $\Phi$, and $\Psi$. The objective for the energy parameters $\Theta$ in minibatch $\mathcal{M}$ is:
\begin{align}
&\hat{\Theta} \!\gets\! 
\argmin_{\Theta}\!\!\!\!\sum_{\langle \x_i, \y_i\rangle\in\mathcal{M}} \nonumber \\ 
&\big[\!\!\cost\!(\canet(\x_i), \y_i) \!-\! E_{\Theta}(\x_i,\canet(\x_i)) \!+\! E_{\Theta}(\x_i, \y_i)\big]_{\! +} \nonumber \\ 
& + \lambda \big[\!-\!\!E_{\Theta}(\x_i,\infnet(\x_i)) \!+\! E_{\Theta}(\x_i, \y_i)\big]_{\! +} \nonumber
\end{align}
When we remove 0-truncation (see Sec.~\ref{sec:truncation} for the motivation), the objective for the  
inference network parameters in minibatch $\mathcal{M}$ is:
\begin{align}
\hat{\Psi}, \hat{\Phi} & \gets \argmax_{\Psi,\Phi}\!\!\sum_{\langle \x_i, \y_i\rangle\in\mathcal{M}} 
\!\cost(\canet(\x_i), \y_i) - \nonumber \\  & E_{\Theta}(\x_i,\canet(\x_i))  
 - \lambda E_{\Theta}(\x_i,\infnet(\x_i)) 
\nonumber
\end{align}
\begin{figure*}[t]
\centering
\includegraphics[width=0.97\textwidth]{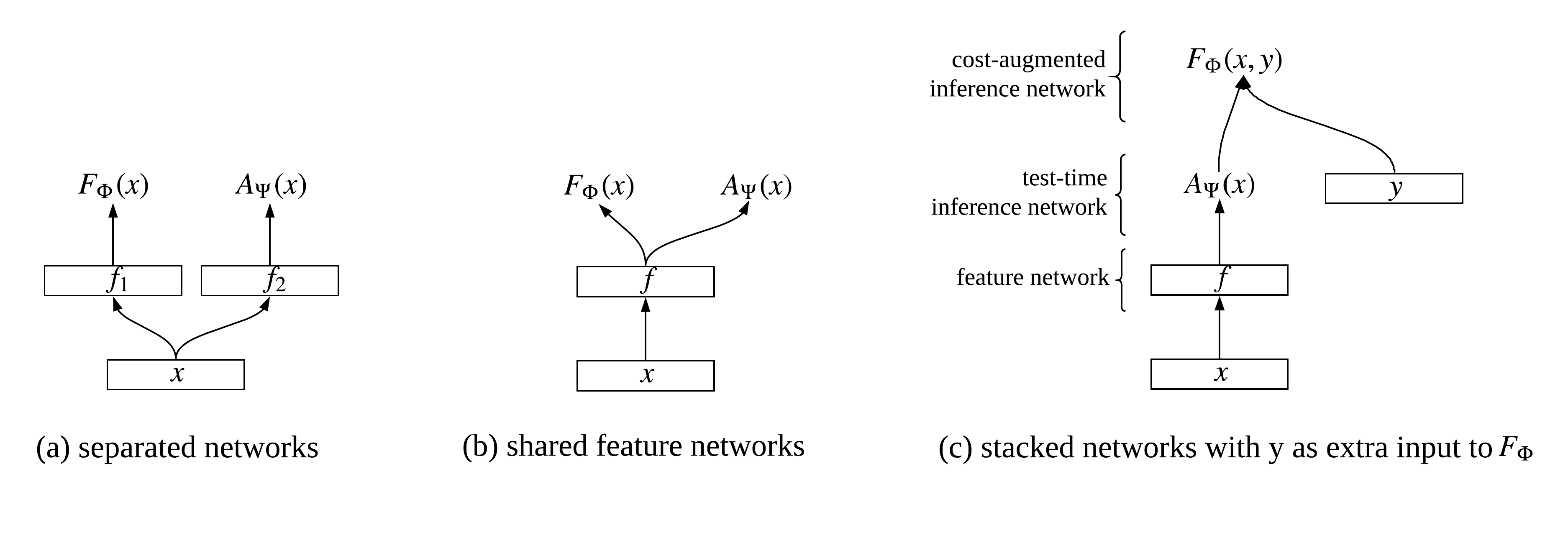}
\caption{
Joint parameterizations for 
cost-augmented inference network $\canet$ and test-time inference network $\infnet$. 
\label{fig:infnets}}
\end{figure*}

\subsection{Joint Parameterizations}\label{sec:joint-param} 

If we were to train independent inference networks $\infnet$ and $\canet$, this new objective could be much slower than the approach of \citet{tu-18}. However, the compound objective offers several natural options for defining joint parameterizations of the two inference networks. 
We consider three options which are visualized in Figure~\ref{fig:infnets} and described below:
\begin{itemizesquish}
  \item \textbf{separated}: $\canet$ and $\infnet$ are two independent networks with their own architectures and parameters as shown in Figure~\ref{fig:infnets}(a).
  \item \textbf{shared}: $\canet$ and $\infnet$ share a ``feature'' network as shown in Figure \ref{fig:infnets}(b). We consider this option because both $\canet$ and $\infnet$ are trained to produce output labels with low energy. However $\canet$ also needs to produce output labels with high cost $\cost$ (i.e., far from the gold standard). 
  \item \textbf{stacked}: the cost-augmented network $\canet$ is a function of the output of the test-time network $\infnet$ \emph{and} the gold standard output $\y$. 
  That is, $\canet (\x,\y)=q(\infnet(\x),\y)$ where $q$ is a parameterized function. This is depicted in Figure~\ref{fig:infnets}(c). 
  We block the gradient at $\infnet$ when updating $\Phi$. 
\end{itemizesquish}

For the $q$ function in the stacked option, we use an affine transformation on the concatenation of the inference network label distribution and the gold standard one-hot vector. That is, denoting the vector at position $t$ of the cost-augmented network output by $\canet(\x, \y)_t$, we have: 
  \begin{align}
   \canet(\x, \y)_t = \softmax(W_q[\infnet(\x)_t;\y(t)]+b_q) \nonumber 
  \end{align}
where semicolon (;) is vertical concatenation, $\y(t)$ (position $t$ of $\y$) is an $L$-dimensional one-hot vector, $\infnet(\x)_t$ is the vector at position $t$ of $\infnet(\x)$, 
$W_q$ is an $L \times 2L$ matrix, and $b_q$ is a bias. 
One motivation for these parameterizations is to reduce the total number of parameters in the procedure. Generally, the number of parameters is expected to decrease when moving from separated to shared to stacked.
We will compare the three options empirically in our experiments, in terms of both accuracy and number of parameters.

Another motivation, specifically for the third option, is to distinguish the two inference networks in terms of their learned functionality. With all three parameterizations, the cost-augmented network will be trained to produce an output that differs from the gold standard, due to the presence of the $\cost(\cdot)$ 
term in the combined objective. However, \citet{tu-18} found that the trained cost-augmented network was barely affected by fine-tuning for the test-time inference objective. This suggests that the cost-augmented network was mostly acting as a test-time inference network by the time of convergence. With the stacked parameterization, however, we explicitly provide the gold standard $\y$ to the cost-augmented network, permitting it to learn to change the predictions of the test-time network in appropriate ways to improve the energy function.

%% file: new_tech.tex
\section{Training Stability and Effectiveness}\label{sec:techniques}

We now discuss several methods that simplify and stabilize training SPENs with inference networks. When describing them, we will illustrate their impact by showing training trajectories for the Twitter part-of-speech tagging task. 
\begin{figure*}[t]
  \centering
  \begin{subfigure}[b]{0.35\linewidth}
    \includegraphics[width=\linewidth]{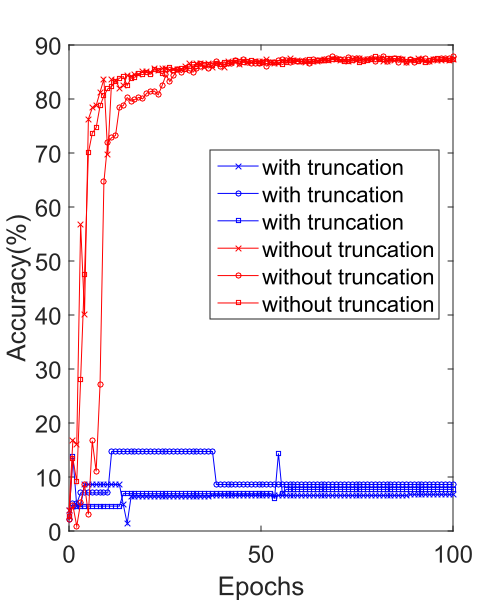}
    \caption{Truncating at 0 (without CE).}
  \end{subfigure}\quad\quad\quad \quad\quad
    \begin{subfigure}[b]{0.35\linewidth}
    \includegraphics[width=\linewidth]{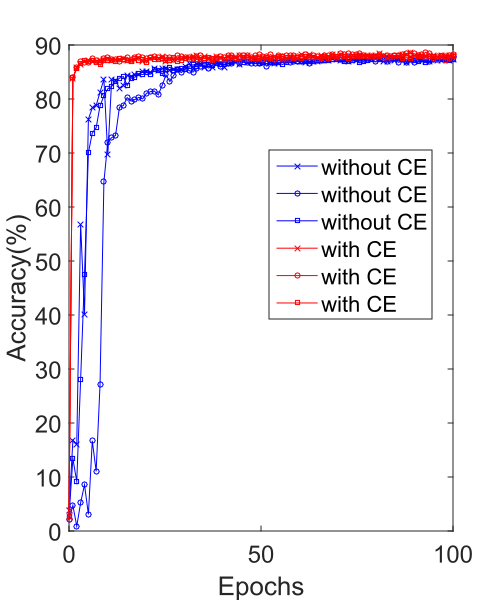}
    \caption{Adding CE loss (without truncation).}
  \end{subfigure}
  \caption{Part-of-speech tagging training trajectories. The three curves in each setting correspond to different random seeds. 
  (a) Without the local CE loss, training fails when using zero truncation. (b) The CE loss reduces the number of epochs for training. 
  \citet{tu-18} used zero truncation and CE during training. 
  \label{fig:trunc}
  }

\end{figure*}

\subsection{Removing Zero Truncation} 
\label{sec:truncation}

\citet{tu-18} used the following objective for the cost-augmented inference network (maximizing it with respect to $\Phi$): $l_0 =$
\begin{align}
[\cost(\canet(\x), \y) - E_{\Theta}(\x,\canet(\x)) + E_{\Theta}(\x, \y)]_{+} \nonumber
\end{align}
\noindent where $[h]_+ = \max(0,h)$. 
However,  
there are two potential reasons why $l_0$ will equal zero and trigger no gradient update. First, $E_\Theta$ (the energy function, corresponding to the  discriminator in a GAN) may already be well-trained, and it can easily separate the gold standard output from the cost-augmented inference network output. Second, the cost-augmented inference network (corresponding to the generator in a GAN) could be so poorly trained that the energy of its output is very large, leading the margin constraints to be satisfied and $l_0$ to be zero. 

In standard margin-rescaled max-margin learning in structured prediction~\citep{m3,ssvm}, the cost-augmented inference step is performed exactly (or approximately with reasonable guarantee of effectiveness), ensuring that when $l_0$ is zero, the energy parameters are well trained. However, in our case, $l_0$ may be zero simply because the cost-augmented inference network is undertrained, which will be the case early in training. 
Then, when using zero truncation, the gradient of the inference network parameters will be 0. This is likely why \citet{tu-18} found it important to add several stabilization terms to the $l_0$ objective.  
We find that by instead removing the truncation, learning stabilizes and becomes less dependent on these additional terms. Note that we retain the truncation at zero when updating the energy parameters $\Theta$. 

As shown in Figure~\ref{fig:trunc}(a), without any stabilization terms and with truncation, the inference network will barely move from its starting point and learning fails overall. 
However, without truncation, the inference network can work well even without any stabilization terms.

\begin{figure*}[th]
  \centering
  \begin{subfigure}[b]{0.3\linewidth}
    \includegraphics[width=\linewidth]{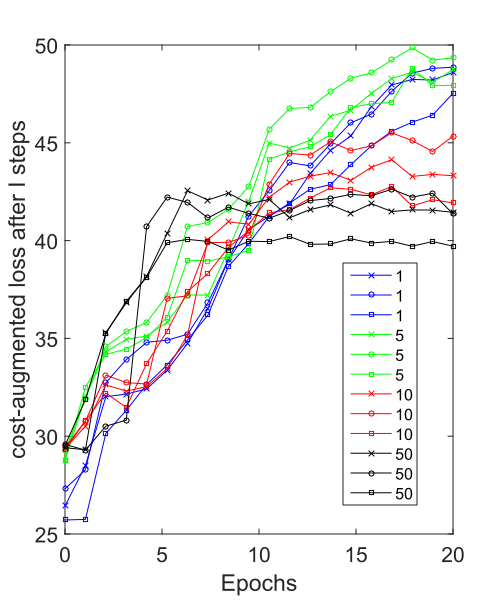}
    \caption{cost-augmented loss $l_1$}
  \end{subfigure}
  \quad\quad\quad
    \begin{subfigure}[b]{0.3\linewidth}
    \includegraphics[width=\linewidth]{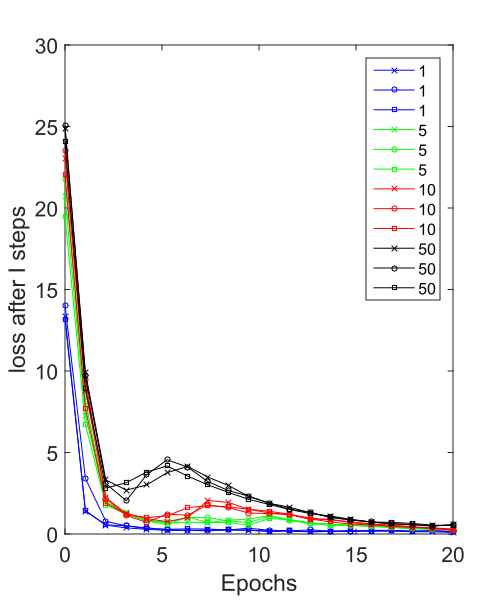}
    \caption{margin-rescaled loss $l_0$}
  \end{subfigure}
  \\
  \begin{subfigure}[b]{0.3\linewidth}
    \includegraphics[width=\linewidth]{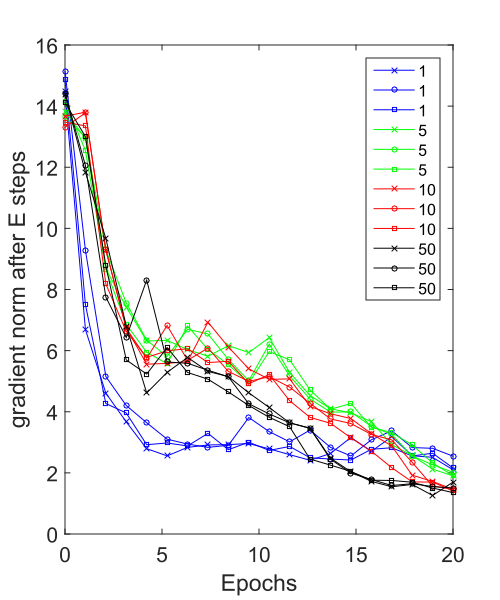}
    \caption{gradient norm of $\Theta$}
  \end{subfigure}
  \quad\quad\quad
  \begin{subfigure}[b]{0.3\linewidth}
    \includegraphics[width=\linewidth]{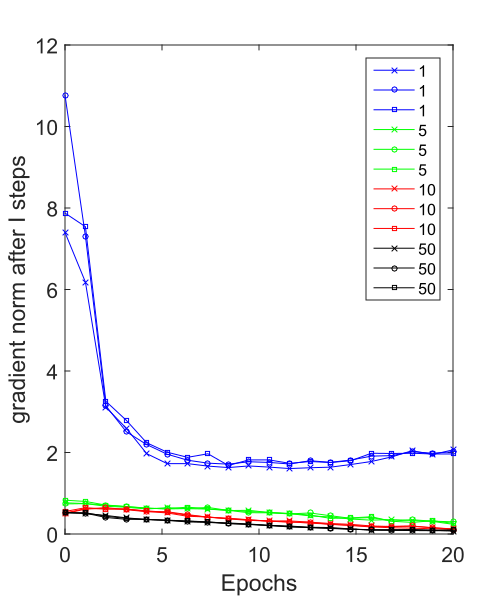}
    \caption{gradient norm of $\Psi$}
  \end{subfigure}
\vspace{-0.1cm}
  \caption{
  POS training trajectories with different numbers of I steps. The three curves in each setting correspond to different random seeds. 
  (a)  cost-augmented loss after I steps; (b) margin-rescaled hinge loss after I steps;  
  (c) gradient norm of energy function parameters after E steps; (d) gradient norm of test-time inference network parameters after I steps. 
  \label{fig:setting_mg}}
\vspace{-0.25cm}
\end{figure*}

\subsection{Local Cross Entropy (CE) Loss}
\label{sec:ce}
\citet{tu-18} proposed adding a local cross entropy (CE) loss, which is the sum of the label cross entropy losses over all positions in the sequence, 
to stabilize inference network training. 
We similarly find this term to help speed up convergence and improve accuracy. 
Figure~\ref{fig:trunc}(b) shows faster convergence to high accuracy when adding the local CE term. See Section~\ref{sec: objective_results} for more details. 

\subsection{Multiple Inference Network Update Steps}
When training SPENs with inference networks, the inference network parameters are nested within the energy function.
We found that the gradient components of the inference network parameters consequently have smaller absolute values than those of the energy function parameters. 
So, we alternate between $k\geq 1$ steps of optimizing the inference network parameters (``I steps'') and one step of optimizing the energy function parameters (``E steps''). We find this strategy especially helpful when using complex inference network architectures. 

To analyze, we compute the 
cost-augmented loss $l_1 = \cost(\canet(\x), \y) - E_{\Theta}(\x,\canet(\x))$ 
and the margin-rescaled hinge loss $l_0 = [\cost(\canet(\x), \y) - E_{\Theta}(\x,\canet(\x)) + E_{\Theta}(\x, \y)]_{+}$ 
averaged over all training pairs $(\x, \y)$ after each set of I steps. 
The I steps update $\Psi$ and $\Phi$ to maximize these  losses. Meanwhile the E steps update $\Theta$ to minimize these losses. 
Figs.~\ref{fig:setting_mg}(a) and (b) show $l_1$ and $l_0$  
during training for different numbers ($k$) of I steps for every one E step. 
Fig.~\ref{fig:setting_mg}(c) shows the norm of the energy parameters after the E steps, 
and 
Fig.~\ref{fig:setting_mg}(d) shows the norm of $\frac{\partial E_{\Theta}(\x, \infnet)}{\partial \Psi}$ after the I steps. 

With $k=1$, the setting used by \citet{tu-18}, the inference network lags behind the energy, making the energy parameter updates very small, as shown by the small norms in Fig.~\ref{fig:setting_mg}(c). The inference network  gradient norm (Fig.~\ref{fig:setting_mg}(d)) remains high, indicating underfitting. 
However, increasing $k$ too much also harms learning, as evidenced by the ``plateau'' effect in the $l_1$ curves for $k=50$; this indicates that the energy function is lagging behind the  inference network. 
Using $k=5$ leads to more of a balance between $l_1$ and $l_0$ and gradient norms that are mostly decreasing during training. We treat $k$ as a hyperparameter that is tuned in our experiments. 

There is a potential connection between our use of multiple I steps and a similar procedure used in GANs \citep{goodfellow2014generative}. In the GAN objective, the discriminator $D$ is updated in the inner loop, and they alternate between multiple update steps for $D$ and one update step for $G$. 
In this section, we similarly found benefit from multiple steps of inner loop optimization for every step of the outer loop. However, the analogy is limited, since GAN training involves sampling noise vectors and using them to generate data, while there are no noise vectors or explicitly-generated samples in our framework. 

%% file: energy.tex
\section{Energies for Sequence Labeling}
\label{sec:tlm-training}
For our sequence labeling experiments in this paper, the input $\x$ is a length-$T$ sequence of tokens, and the output $\y$ is a sequence of labels of length $T$. 
We use $\y_t$ to denote the output label at position $t$, where $\y_t$ is a vector of length $L$ (the number of labels in the label set) and where $y_{t,j}$ is the $j$th entry of the vector $\y_t$. In the original output space $\yspace(\x)$, $y_{t,j}$ is 1 for a single $j$ and 0 for all others. In the relaxed output space $\relyspace(\x)$, $y_{t,j}$  can be interpreted as the probability of the $t$th position being labeled with label $j$. 
We then use the following energy for sequence labeling~\citep{tu-18}: 
\begin{align}
\label{eqn:energy-sequence-labeling}
E_{\Theta}(\x, \y) &= -\Bigg(\sum_{t=1}^T  \sum_{j=1}^L y_{t,j}\! \left(U_j^\top b(\x,t)\right) \nonumber \\
& + \sum_{t=1}^T \y_{t-1}^\top W \y_{t}\Bigg) 
\end{align}
where $U_{j}\in\mathbb{R}^d$ is a parameter vector for label $j$  
and the parameter matrix $W\in\mathbb{R}^{L \times L}$ contains label-pair parameters. 
Also, $b(\x,t)\in\mathbb{R}^d$ denotes the ``input feature vector'' for position $t$. 
We define $b$ to be the $d$-dimensional BiLSTM~\citep{Hochreiter:1997:LSM:1246443.1246450} hidden vector at $t$.
The full set of energy parameters $\Theta$ includes the $U_j$ vectors, $W$, and the parameters of the BiLSTM.

\paragraph{Global Energies for Sequence Labeling.}
In addition to new training strategies, we also experiment with several global energy terms for sequence labeling. 
Eq.~(\ref{eqn:energy-sequence-labeling}) shows the base energy, and to capture long-distance  dependencies, we include global energy (GE) terms in the form of Eq.~(\ref{eqn:basic-tlm}). 

We use $h$ to denote an LSTM tag language model (TLM) that takes a sequence of labels as input and returns a distribution over next labels. 
We define $\overline{\vy}_t = {h}(\vy_0,\dots,\vy_{t-1})$ to be the distribution given the preceding label vectors (under a LSTM language model). Then, the energy term is:
\begin{align}
E^{\mathrm{TLM}}(\y) = - \sum_{t=1}^{T+1} \log \left(\vy_t^\top \overline{\vy}_t\right)
\label{eqn:basic-tlm}
\end{align}
\noindent 
where $\y_0$ is the start-of-sequence symbol and  $\y_{T+1}$ is the end-of-sequence symbol. 
This energy returns the negative log-likelihood under the TLM of the candidate output $\y$. 
\citet{tu-18} pretrained their $h$ on a large,  automatically-tagged corpus and fixed its parameters when optimizing $\Theta$. 
Our approach has one critical difference. We instead \textit{do not} pretrain $h$, and its parameters are learned when optimizing $\Theta$. We show that even without pretraining, our global energy terms are still able to capture useful additional information.

We also propose new global energy terms. Define $\overline{\vy}_t = {h}(\vy_0,\dots,\vy_{t-1})$ where $h$ is an LSTM TLM that takes a sequence of labels as input and returns a distribution over next labels. 
First, we add a TLM in the backward direction (denoted $\overline{\vy}'_t$ analogously to the forward TLM). 
Second, we include words as additional inputs to forward and backward TLMs.  
We define $\widetilde{\vy}_t = {g}(\vx_0,..., \vx_{t-1}, \vy_0,...,\vy_{t-1})$ where $g$ is a forward LSTM TLM.
We define the backward version similarly (denoted $\widetilde{\vy}'_t$). The global energy is therefore
\begin{align}
E^{\mathrm{GE}}(\y) &= - \sum_{t=1}^{T+1} 
\log (\vy_t^\top \overline{\vy}_t) +  \log (\vy_t^\top \overline{\vy}'_t) \nonumber \\
& + \gamma\big(\log (\vy_t^\top \widetilde{\vy}_t)  + \log (\vy_t^\top \widetilde{\vy}'_t)\big) 
\label{eqn:stronger-tlm}
\end{align}
Here $\gamma$ is a hyperparameter that is tuned. We experiment with three settings for the global energy: GE(a): forward TLM as in \citet{tu-18}; GE(b): forward and backward TLMs  ($\gamma = 0$); GE(c): all four TLMs in Eq.~(\ref{eqn:stronger-tlm}).

%% file: ExperimentSetup.tex
\section{Experimental Setup}
\label{sec:expsetup}

We consider two sequence labeling tasks: Twitter part-of-speech (POS) tagging~\citep{gimpel-etal-2011-part} and named entity recognition (NER;~\citealp{conll20003}).

\paragraph{Twitter Part-of-Speech (POS) Tagging.}

We use the 
Twitter POS data from \citet{gimpel-etal-2011-part} and \citet{owoputi-etal-2013-improved} which contain 25 tags. We use 100-dimensional skip-gram \citep{mikolov2013distributed} embeddings from \citet{tu-17-long}. 
Like \citet{tu-18}, we use a BiLSTM to compute the input feature vector for each position, using hidden size 100. We also use BiLSTMs for the inference networks. The output of the inference network is a softmax function, so the inference network will produce a distribution over labels at each position. The $\Delta$ is L1 distance. We train the inference network using stochastic gradient descent (SGD) with momentum and train the energy parameters using Adam~\citep{adam}. We also explore training the inference network using Adam when not using the local CE loss.\footnote{We find that Adam works better than SGD when training the inference network without the local cross entropy term.} In experiments with the local CE term, its weight is set to 1.

\paragraph{Named Entity Recognition (NER).}

We use the CoNLL 2003 English dataset \citep{conll20003}. We use the BIOES tagging scheme, following previous work \citep{Ratinov:2009},  resulting in 17 NER labels. We use 100-dimensional pretrained GloVe embeddings \citep{pennington-etal-2014-glove}. 
The task is evaluated using F1 score computed with the \texttt{conlleval} script.   
The architectures for the feature networks in the energy function and inference networks are all BiLSTMs. The architectures for tag language models are LSTMs. 
We use a dropout \texttt{keep-prob} of 0.7 for all LSTM cells. The hidden size for all LSTMs is 128. We use Adam~\citep{adam} and do early stopping on the development set. We use a learning rate of $5\cdot 10^{-4}$. Similar to above, the weight for the CE term is set to 1. 

We consider three NER modeling configurations. \textbf{NER} uses only words as input and pretrained, fixed GloVe embeddings. \textbf{NER+} uses words, the case of the first letter, POS tags, and chunk labels, as well as pretrained GloVe embeddings with fine-tuning. \textbf{NER++} includes everything in \textbf{NER+} as well as character-based word representations obtained using a convolutional network over the character sequence in each word.  Unless otherwise indicated, our SPENs use the energy in Eq.~(\ref{eqn:energy-sequence-labeling}).

%% file: experiments.tex
\section{Results and Analysis}\label{sec:results}

\label{sec: objective_results}

\begin{table}[t]
\setlength{\tabcolsep}{5pt}
\small
\begin{center}
\begin{tabular}{ccc|c||c||c|}
\cline{4-6}
 &zero&& \multicolumn{1}{c||}{POS}
& \multicolumn{1}{c||}{NER}
& \multicolumn{1}{c|}{NER+}
\\ \cline{4-6}
loss &trunc.&CE & acc (\%) & F1 (\%) & F1 (\%)\\ 
\hline
\multicolumn{1}{|c}{} & yes & no & 13.9  &  3.91  & 3.91\\
\multicolumn{1}{|c}{margin-} & no & no & 87.9 & 85.1 & 88.6\\
\multicolumn{1}{|c}{rescaled} & yes & yes & 89.4{{\makebox[0pt][l]{*}}} &  85.2{{\makebox[0pt][l]{*}}} &  89.5{{\makebox[0pt][l]{*}}} \\
\multicolumn{1}{|c}{} & no & yes & 89.4 & 85.2 & 89.5\\
\hline

\multicolumn{1}{|c}{\multirow{2}{*}{perceptron}} & no & no & 88.2  & 84.0 & 88.1\\
\multicolumn{1}{|c}{} & no & yes & 88.6 & 84.7 & 89.0 \\
\hline
\end{tabular}
\end{center}
\caption{Test results for POS and NER for several SPEN configurations. Results with * correspond to the setting of \citet{tu-18}. The inference network architecture is a one-layer BiLSTM. \label{table:results2}} 
\end{table}

\begin{table*}[th]
\setlength{\tabcolsep}{4pt}
\small
\begin{center}
\begin{tabular}{lc|c|c|c|c||c|c|c|c||c|}
\cline{3-11}
&& \multicolumn{4}{c||}{POS}
& \multicolumn{4}{c||}{NER}
& \multicolumn{1}{c|}{NER+}
\\ 
 && \multicolumn{1}{c}{acc. (\%)} & \multicolumn{1}{c}{$|T|$} &    \multicolumn{1}{c}{$|I|$}  &\multicolumn{1}{c||}{speed} & \multicolumn{1}{c}{F1 (\%)} & \multicolumn{1}{c}{$|T|$} &  \multicolumn{1}{c}{$|I|$} & \multicolumn{1}{c||}{speed}  & F1 (\%) \\
\hline
\multicolumn{2}{|l|}{BiLSTM}   & 88.8 & 166K  &166K & -- & 84.9 & 239K & 239K & -- &  89.3 \\
\hline
\\ 
\multicolumn{9}{l}{\textbf{SPENs with inference networks \citep{tu-18}:}} \\
\hline
\multicolumn{2}{|l|}{margin-rescaled} & 89.4 & 333K  & 166K & -- & 85.2 & 479K &239K & -- &89.5 \\

\multicolumn{2}{|l|}{perceptron} & 88.6 & 333K &166K & -- & 84.4 & 479K & 239K & -- & 89.0 \\
\hline
\\
\multicolumn{11}{l}{\textbf{SPENs with inference networks, compound objective, CE, no zero truncation (this paper):}} \\
\hline
\multicolumn{2}{|l|}{separated} & 89.7 & 500K & 166K & 66 &  85.0 & 719K & 239K & 32 & 89.8 \\
\multicolumn{2}{|l|}{shared} & 89.8 & 339K & 166K & 78 & 85.6 & 485K & 239K & 38 & 90.1 \\
\multicolumn{2}{|l|}{\bf{stacked}} & \bf{89.8} & \bf{335K} & \bf{166K} &  \bf{92} & \bf{85.6} & \bf{481K} & \bf{239K} & \bf{46} & \bf{90.1} \\
\hline
\end{tabular}
\end{center}
\caption{Test results for POS and NER. 
$|T|$ is the number of trained parameters; $|I|$ is the number of parameters needed during inference. 
Training speeds (examples/second) are shown for joint parameterizations to compare them in terms of efficiency. Best setting (best performance with fewest parameters and fastest training) is in bold. 
\label{table:results1}}
\end{table*}

\paragraph{Effect of Zero Truncation and Local CE Loss.}
Table~\ref{table:results2} shows results for zero truncation and the local CE term. 
Training fails for both tasks when using zero truncation without CE. Removing truncation makes learning succeed and leads to effective models even without using CE. 
However, when using the local CE term, truncation has little effect on performance. The importance of CE in prior work \citep{tu-18} is likely due to the fact that truncation was being used. 

The local CE term is useful for both tasks, though it appears more helpful for tagging.\footnote{We found the local CE term to be useful for both the cost-augmented and test-time inference networks during training.} 
This may be because POS tagging is a more local task. 
Regardless, for both tasks, as shown in Section~\ref{sec:ce}, the inclusion of the CE term speeds convergence and improves training stability. 
For example, on NER, using the CE term reduces the number of epochs chosen by early stopping from $\sim$100 to $\sim$25. For POS, using the CE term reduces the number of epochs from $\sim$150 to $\sim$60.

\paragraph{Effect of Compound Objective and Joint Parameterizations.}
The compound objective is the sum of the margin-rescaled and perceptron losses, and outperforms them both (see Table \ref{table:results1}). 
Across all tasks, the shared and stacked parameterizations are more accurate than the previous objectives. For the separated parameterization, the performance drops slightly for NER, likely due to the larger number of parameters. 
The shared and stacked options have fewer parameters to train than the separated option, and the stacked version processes examples at the fastest rate during training.

\begin{table}[t]
\small
\centering
\begin{tabular}{lc|c||c|}
\cline{3-4}
& & \multicolumn{1}{c||}{POS}
& \multicolumn{1}{c|}{NER}
\\ \cline{3-4}
   &&  $  \infnet - \canet  $ &  $ \infnet - \canet  $ \\
\hline
\multicolumn{2}{|l|}{margin-rescaled} & 0.2 & 0\\
\hline
\multicolumn{1}{|l}{} & separated & 2.2 & 0.4\\
\multicolumn{1}{|l}{compound} & shared & 1.9  & 0.5\\
\multicolumn{1}{|l}{} & stacked & \bf{2.6}  & \bf{1.7}\\
\hline
\end{tabular}
\\
\medskip
\begin{tabular}{cc}
test-time ($\infnet$)  & cost-augmented ($\canet$) \\
\hline
 common noun  & proper noun \\
 proper noun & common noun   \\
 common noun  & adjective  \\
 proper noun  & proper noun + possessive\\
 adverb & adjective \\
 preposition & adverb \\
 adverb & preposition \\
 verb &  common noun\\
 adjective & verb \\
\end{tabular}
\caption{Top: differences in accuracy/F1 between test-time inference networks $\infnet$ and cost-augmented networks $\canet$ (on development sets). 
The ``margin-rescaled'' row uses a SPEN with the local CE term and without zero truncation, where $\infnet$ is obtained by fine-tuning $\canet$ as done by \citet{tu-18}. Bottom: most frequent output differences between $\infnet$ and $\canet$ on the development set. 
\label{table:phi-vs-psi}
}
\end{table}

The top part of Table~\ref{table:phi-vs-psi} shows how the performance of the test-time inference network $\infnet$ and the cost-augmented inference network $\canet$ vary when using the new compound objective. 
The differences between $\canet$ and $\infnet$ are larger than in the baseline configuration, showing that the two are learning complementary functionality. 
With the stacked parameterization, the cost-augmented network $\canet$ receives as an additional input the gold standard label sequence, which leads to the largest differences as the cost-augmented network can explicitly favor incorrect labels.\footnote{We also tried a BiLSTM in the final layer of the stacked parameterization but results were similar to the simpler affine architecture, so we only report results for the latter.}

The bottom part of Table~\ref{table:phi-vs-psi} shows qualitative differences between the two inference networks. On the POS development set, we count the differences between the predictions of $\infnet$ and $\canet$ when 
$\infnet$ makes the correct prediction.\footnote{We used the stacked parameterization.} 
$\canet$ tends to output tags that are highly confusable with those output by $\infnet$. For example, it often outputs proper noun when the gold standard is common noun or vice versa. It also captures the 
ambiguities among adverbs, adjectives, and prepositions. 

\paragraph{Global Energies.}
The results are shown in Table \ref{table:tlm_result}.
Adding the backward (b) and word-augmented TLMs (c) improves over using only the forward TLM from \citet{tu-18}. 
With the global energies, our performance is comparable to several strong results (90.94 of \citealp{lample-etal-2016-neural} and 91.37 of \citealp{ma-hovy-2016-end}). However, it is still lower than the state of the art~\citep{akbik-etal-2018-contextual,devlin-etal-2019-bert}, likely due to the lack of 
contextualized embeddings. 
In other work, we proposed and evaluated several other high-order energy terms for sequence labeling using our framework \citep{ArbitraryOrder}.

\begin{table}[th!]
\centering
\small
\begin{tabular}{|m{3cm}|c|c|c|}
\cline{2-4}
\multicolumn{1}{l|}{}  & NER & NER+ & NER++\\ 
\hline
margin-rescaled & 85.2 & 89.5 & 90.2\\
\hline
{compound, stacked, CE, no truncation} & 85.6 & 90.1 & 90.8\\ 
\hline
+ global energy GE(a) & 85.8 & 90.2 & 90.7\\
+ global energy GE(b) & 85.9 & 90.2 & 90.8\\
+ global energy GE(c) &  \bf{86.3} & \bf{90.4} & \bf{91.0}\\
\hline
\end{tabular}
\caption{NER test F1 scores with global energy terms. 
}
\label{table:tlm_result}
\end{table}

%% file: relatedwork.tex
\section{Related Work}

There are several efforts aimed at stabilizing and improving learning in generative adversarial networks (GANs) \citep{goodfellow2014generative,NIPS2016_6125,DBLP:journals/corr/ZhaoML16,Arjovsky2017WassersteinG}. 
Progress in training GANs has come largely from overcoming learning difficulties by modifying loss functions and optimization, and GANs have become more successful and popular as a result. 
Notably, Wasserstein GANs \citep{Arjovsky2017WassersteinG} provided the first convergence measure in GAN training using Wasserstein distance. 
To compute Wasserstein distance, the discriminator uses weight clipping, which 
limits network capacity. 
Weight clipping was subsequently replaced with a gradient norm constraint \citep{NIPS2017_7159}. \citet{miyato2018spectral} proposed a novel weight normalization technique called spectral normalization. These methods may be applicable to the similar optimization problems solved in learning SPENs. 
Another direction may be to explore alternative training objectives for SPENs, such as those that use weaker supervision than complete structures~\citep{N18-2021,NIPS2019_9507,naskar2020energy}.